%% file: 2026_05_07_bioresearcher_arxiv.tex
\documentclass{article}

\usepackage[preprint]{neurips_2026}

\usepackage[utf8]{inputenc} 
\usepackage[T1]{fontenc}    
\usepackage{hyperref}       
\usepackage{url}            
\usepackage{booktabs}       
\usepackage{amsfonts}       
\usepackage{amsmath}
\usepackage{amssymb}
\usepackage{amsthm}

\usepackage{nicefrac}       
\usepackage{microtype}      
\usepackage{xcolor}         
\usepackage{graphicx}
\usepackage{tikz}
\usepackage{pgfplots}
\usepackage{algorithm}
\usepackage{algorithmic}
\usepackage{multirow}
\usepackage{enumitem}
\usepackage{subcaption}
\usepackage{tcolorbox}

\usetikzlibrary{
  arrows.meta,
  positioning,
  shapes.geometric,
  shapes.multipart,
  fit,
  backgrounds,
  calc,
  decorations.pathreplacing,
  patterns
}

\pgfplotsset{compat=1.18}
\graphicspath{{media/}{figs/}}

\newcommand{\modelname}{\textsc{BioResearcher}}
\input{generated/biomarker_e2e_stats}

\title{\modelname: Scenario-Guided Multi-Agent for Translational Medicine}

\author{%
  Remigiusz Kinas  \and
  Joanna Krawczyk \and
  Rafał Powalski \and
  Przemysław Pietrzak \and
  Agnieszka Kowalewska \and
  Krzysztof Kolmus \and
  Maciej Sypetkowski \and
  Łukasz Smoliński \and
  Tomasz Jetka \thanks{Corresponding author: \texttt{tomasz.jetka@ingenix.ai}} \AND
  {$\ $}\\
  Ingenix.AI, Warsaw, Poland \\
}

\begin{document}
\maketitle


\input{sections_4pages/0_abstract}

\input{sections_4pages/1_introduction}

\input{sections_4pages/2_system_architecture}

\input{sections_4pages/3_evaluation}

\input{sections_4pages/4_results}

\input{sections_4pages/7_analysis_discussion}

\newpage
\bibliographystyle{unsrt}
\bibliography{references}

\newpage
\appendix

\input{appendix/architecture}

\input{appendix/tools}

\input{appendix/benchmarks}

\input{appendix/related_work}

\input{appendix/cases}

\input{appendix/licences}

\clearpage

\end{document}

%% file: generated/biomarker_e2e_stats.tex
\providecommand{\BioEtwoPosMeanPct}{74.7}
\providecommand{\BioEtwoPosSdPct}{3.3}
\providecommand{\BioEtwoNegMeanPct}{96.8}
\providecommand{\BioEtwoNegSdPct}{0.2}

%% file: sections_4pages/0_abstract.tex
\begin{abstract}
Translational medicine turns underspecified development goals into evidence synthesis that must combine literature, trials, patents, and quantitative multi-omics analysis while preserving identifiers, uncertainty, and retrievable provenance. General-purpose foundation models and off-the-shelf tool-augmented or multi-agent systems are not built for this: they tend to produce single-shot answers or run open-endedly, and fall short on the auditable, scenario-specific workflows that heterogeneous biomedical sources demand.

This paper introduces \modelname{}, a scenario-guided multi-agent system that maps queries to versioned research playbooks, delegates to specialized subagents over 30+ tools and machine-learning endpoints, mixes structured database access with sandboxed code for genome-scale analyses, and applies claim-level multi-model reconciliation before editorial assembly.

We evaluate \modelname{} across unit-level capabilities, open-ended biomedical reasoning, and end-to-end clinical discovery. It leads evaluated baselines on 109 single-step tests (83.49\% pass rate; 0.892 average score), achieves strong biomedical benchmark performance (89.33\% on BixBench-Verified-50 and the top 0.758 mean score on BaisBench Scientific Discovery), and leads on a 30-query clinical end-to-end benchmark with the highest positive hit rate (\BioEtwoPosMeanPct\%\ $\pm$\ \BioEtwoPosSdPct\%) and negative clear rate (\BioEtwoNegMeanPct\%\ $\pm$\ \BioEtwoNegSdPct\%). These results show broad, competitive performance across unit-level, open-ended, and end-to-end clinical evaluations.

\end{abstract}

%% file: sections_4pages/1_introduction.tex
\section{Introduction}
\label{sec:introduction}

Consider the following, exemplar question posed to a translational research team: \emph{``Given an antibody--drug conjugate (ADC) with a topoisomerase-I (TOP1) primary payload, what is the optimal second payload class for synergistic anti-tumor activity in non-small cell lung cancer (NSCLC)?''}

This question, representative of real strategic decisions in pharmaceutical R\&D, is operationally hard for several reasons.
First, the query is \textbf{entity-ambiguous}: ``Topo-I,'' ``topo1,'' ``topoisomerase I,'' and ``TOP1'' all denote the same protein target (ChEMBL ID CHEMBL1781, Ensembl ENSG00000198900), yet different databases index them under different names.
At the same time, \textbf{ontology mismatches} complicate cross-source integration: disease identifiers vary across EFO, DOID, MONDO, and TCGA conventions; compound identifiers differ between ChEMBL, and PubChem while accurate drug's target mapping is noisy and depends on application.

Second, relevant evidence is \textbf{scattered across heterogeneous sources}: PubMed abstracts, ClinicalTrials.gov records, Google Patents filings, and DepMap multi-omics datasets, each with distinct identifiers, access patterns, and biological
  context. Additional sources such as research documents and ASCO, AACR, or ESMO abstracts could be integrated where licensing permits.

  Third, reasoning spans interacting biological scales and high-dimensional measurements, requiring both qualitative synthesis and quantitative analysis. Fourth, outputs must be auditable dossiers: ranked hypotheses with mechanistic rationale and
  provenance. We present \modelname{}, a scenario-guided multi-agent system that normalizes entities, routes questions to translational playbooks, delegates retrieval and CodeAct analyses, and reconciles evidence into provenance-preserving reports.

Our contributions are:
\begin{enumerate}[label=(\alph*),leftmargin=*,nosep]
  \item \textbf{A scenario-guided agent architecture for translational medicine} that frames the problem as scientific assistance for daily translational workflows: translating broad questions into explicit research plans, combining qualitative and quantitative evidence, and contextualizing outputs for clinical development (Section~\ref{sec:architecture}).
  \item \textbf{A layered evidence-synthesis stack with multi-model reconciliation} that integrates ontology-aware translation, qualitative knowledge synthesis, quantitative dataset analysis, and autonomous CodeAct-style computation under a single orchestrator, and reconciles outputs via structured claim-level debate---claim extraction, cross-model grouping, multi-round argumentation, and quantitative consensus detection---for auditable long-form biomedical report synthesis (Section~\ref{sec:architecture}).
  \item \textbf{A realistic multi-level evaluation protocol} that separates component verification from hierarchical reasoning tests, autonomous custom-analysis tasks, and end-to-end expert assessment on realistic translational questions, together with ablations against direct LLM generation (Section~\ref{sec:evaluation}).
\end{enumerate}

%% file: sections_4pages/2_system_architecture.tex
\section{System Architecture}
\label{sec:architecture}

\textsc{BioResearcher} is a scenario-guided multi-agent system for translational evidence synthesis.
Natural-language questions are often underspecified.
Outputs are auditable dossiers with normalized entities, heterogeneous evidence, ranked hypotheses, mechanistic links, and retrievable provenance (e.g., PMIDs, NCT IDs, and patent numbers).

The architecture separates method selection, evidence acquisition, and reconciliation (\autoref{fig:architecture}, Appendix~\ref{app:architecture}). A master orchestrator selects a versioned scenario playbook, decomposes the query, and delegates to
  specialized, state-isolated subagents that publish provenanced artifacts. A reconciliation agent drafts, compares, and resolves claims before a light editorial pass. Subagents use tool loops, iterative synthesis, cross-provider fan-out, and sandboxed
  code-mediated analysis; implementation details and tool coverage are provided in Appendices~\ref{app:architecture} and~\ref{app:tools}.

%% file: sections_4pages/3_evaluation.tex
\section{Evaluation}
\label{sec:evaluation}

To rigorously evaluate capability of \modelname, we designed a benchmark suite at three complementary conceptual levels: single-step tests, clinical reasoning \& autonomous analysis benchmark, and clinical end-to-end benchmark.

\subsection{Single-Step Tests}

We assembled a 109-question benchmark that verifies each system layer in isolation: L1 ontology and entity resolution, L2 qualitative retrieval, and L3 quantitative computation. That enables to verify how the system performs with accurately answering questions about data and facts across drug development aspects at every modality scale.
The purpose of this benchmark is to see if the agent can perform simple factual checks and calculations.
One may consider this a sanity check to see if the agent is able to perform basic tasks: crucial before evaluating the agent on more complex queries.
The questions check e.g. the ability to resolve entity IDs, retrieve a PubMed publication, and process quantitative data.
Each test question asks about a functionality of one sub-agent.
Therefore, the tests do not only check the ability of the agent to perform single-step reasoning, but also the ability of the agent to route a simple question to the correct sub-agent.
We use the single-step tests to compare the performance of \modelname{}, GPT-5.4-mini, GPT-5.5, and CellType Agent.
The detailed benchmark design (subset definitions, test-case format) and evaluation protocol (LLM-as-a-judge configuration, rubric, scoring metric) are given in Appendix~\ref{app:benchmark}.

\subsection{Quantitative Reasoning \& Autonomous Analysis Benchmark}
\label{sec:qr_autonomous_benchmark}

We evaluate \textsc{\modelname{}}'s ability to analyze custom datasets and propose dedicated autonomous analyses when the required computation is not fully specified in advance, using two complementary benchmarks: \textit{BixBench}~\cite{mitchener_bixbench_2025} and \textit{BaisBench}~(Scientific Discovery)~\cite{luo_benchmarking_2026}.
See Appendix~\ref{app:benchmark} for benchmark descriptions, task setup, and scoring details.

\subsection{Clinical End-to-End Biomarkers Benchmark.}
To evaluate end-to-end performance on translational-medicine reasoning, we
constructed a benchmark of 30 expert-curated queries that pair a drug,
mechanism of action (MOA), or indication with a clinically or preclinically
validated biomarker. Items are stratified across seven tiers of evidential
maturity, ranging from biomarkers approved alongside marketed therapies to
prognostic biomarkers and synthetic-lethality hypotheses still under
investigation.

Each query is paired with (i) an expert-authored positive
answer drawn from regulatory labels, pivotal trials, or peer-reviewed
literature, and (ii) a plausible negative control selected from adjacent
biology to probe discrimination rather than mere recall.
We report two metrics and then averaged across all benchmark items: the \emph{positive sample rate} (PSR), i.e.\ the fraction of items whose positive answer $p_i$ appears in the system's Top-10 list $L_i$, and the \emph{negative sample rate} (NSR), i.e.\ the fraction of items whose negative control $n_i$ is correctly excluded from $L_i$ (formal definitions in Appendix~\ref{app:benchmark}).
The whole benchmark is run three times per system, and the reported PSR and NSR are averaged.

%% file: sections_4pages/4_results.tex
\section{Results}
\label{sec:results}

\subsection{Single-Step Tests}
Evaluation results for \modelname{}, GPT-5.4-mini, GPT-5.5, and CellType Agent are shown in \autoref{tab:single_step_results}.
As expected, GPT-5.4-mini has the lowest pass rates and average metric score across all layers and subsets.
GPT-5.5 has signifficantly higher performance than GPT-5.4-mini, in all metrics.
CellType Agent has the highest pass rate in layer L2 (Qualitative Synthesis).
\modelname{} with GPT-5.4 as core model achieved the highest average metric score in layers L1 (Ontology \& Entity) and L3 (Quantitative Analysis),
and the highest total pass rate and average metric score overall.
\begin{table}[!ht]
\caption{Single-step test results, averaged over 3 runs.
The first four columns report pass rates,
defined as the percentage of cases in the top rubric band
(mean~$\pm$~std over 3 runs) per layer and overall.
The last two columns report the judge’s average metric score and the average token usage on the entire evaluation set (mean~$\pm$~std over 3 runs).
Model suffixes indicate reasoning effort, and \modelname{} variants
indicate the core model used. \modelname{} (mini) stands for \modelname{} with core model
GPT-5.4-mini, \modelname{} (low) stands for \modelname{} with core model GPT-5.4 with reasoning effort set to “low”.}
\label{tab:single_step_results}
\centering
\small
\begin{tabular}{@{}lcccccc@{}}
\toprule
\textbf{System} & \textbf{L1 (\%)} & \textbf{L2 (\%)} & \textbf{L3 (\%)} & \textbf{Total (\%)} & \textbf{Avg score}  & \textbf{Token usage} \\
\midrule
GPT-5.4-mini & 26.26 $\pm$ 6.31 & 17.12 $\pm$ 4.13 & 11.11 $\pm$ 3.92 & 17.74 $\pm$ 2.80 & 0.316 $\pm$ 0.022 & 19167 $\pm$ 3948 \\
GPT-5.4 (low) & 40.85 $\pm$ 3.94 & 26.13 $\pm$ 3.12 & 20.51 $\pm$ 0.00 & 28.53 $\pm$ 1.84 & 0.448 $\pm$ 0.008 & 61824 $\pm$ 2341 \\
GPT-5.5 (medium) & 65.66 $\pm$ 1.75 & 37.84 $\pm$ 4.68 & 28.21 $\pm$ 5.13 & 42.81 $\pm$ 2.12 & 0.585 $\pm$ 0.009 & 221829 $\pm$ 7968 \\
GPT-5.5 (high) & 67.68 $\pm$ 1.75 & 40.04 $\pm$ 4.67 & 30.77 $\pm$ 2.56 & 45.09 $\pm$ 0.24 & 0.591 $\pm$ 0.005 & 447181 $\pm$ 20637 \\
Gemini 3.1 Pro & 47.38 $\pm$ 3.24 & 27.17 $\pm$ 1.21 & 12.82 $\pm$ 0.00 & 27.94 $\pm$ 1.10 & 0.481 $\pm$ 0.009 & 176974 $\pm$ 7500 \\
CellType & 81.82 $\pm$ 3.03 & \textbf{77.48 $\pm$ 1.56} & 46.60 $\pm$ 7.32 & 67.80 $\pm$ 1.70 & 0.731 $\pm$ 0.003 & 51953004 $\pm$ 931350 \\
\textbf{\modelname{}} (mini) & 77.78 $\pm$ 1.75 & 74.77 $\pm$ 3.12 & 91.45 $\pm$ 1.48 & 81.65 $\pm$ 0.92 & 0.889 $\pm$ 0.015 & 2975729 $\pm$ 8001 \\
\textbf{\modelname{}} (low) & \textbf{84.85 $\pm$ 3.03} & 68.47 $\pm$ 3.12 & \textbf{96.58 $\pm$ 1.48} & \textbf{83.49 $\pm$ 1.59} & \textbf{0.892 $\pm$ 0.004} & 2646126 $\pm$ 29309 \\
\bottomrule
\end{tabular}
\end{table}

\subsection{Quantitative Reasoning \& Autonomous Analysis Benchmark}

\paragraph{BixBench.}
We report BixBench-style open-answer accuracy on \texttt{phylobio/BixBench-Verified-50}~\cite{mitchener_bixbench_2025} (\autoref{tab:bixbench_results}) for the full 50-question set and a 22-question \emph{human} subset of questions that concern human biology.
For fairness, package-augmented Claude Code variants were provided the same data-analysis package list available to \modelname{}, ensuring minimal execution context.

\paragraph{BaisBench.}
Mean~$S_{\mathrm{SD}}$ on the Scientific Discovery (BAIS-SD) track of BaisBench~\cite{luo_benchmarking_2026} is reported in \autoref{tab:baisbench_results}.
Task setup and scoring are specified in \autoref{sec:qr_autonomous_benchmark}.

\begin{table}[!ht]
\setlength{\abovecaptionskip}{4pt}
\setlength{\belowcaptionskip}{2pt}
\caption{Quantitative reasoning and autonomous analysis benchmark results.
BixBench-Verified-50~\cite{mitchener_bixbench_2025} reports open-answer accuracy (\%, mean~$\pm$~std over 3 runs) on the full set (50 questions) and a 22-question subset of questions about human biology.
BaisBench Scientific Discovery (BAIS-SD)~\cite{luo_benchmarking_2026} reports mean $S_{\mathrm{SD}}$ on the 193-question SD track.
\modelname{} uses core model GPT-5.4 with reasoning effort set to ``medium''; No value is bolded since all results are statistically indistinguishable from each other.
CellType and Claude Code are powered by Claude Opus 4.7.}
\label{tab:bixbench_results}
\label{tab:baisbench_results}
\centering
\small
\begin{tabular}{@{}lccc@{}}
\toprule
\textbf{System} & \multicolumn{2}{c}{\textbf{BixBench acc. (\%)}} & \textbf{BAIS-SD} \\
\cmidrule(lr){2-3}
 & \textbf{Full} & \textbf{Human} & \textbf{Mean score} \\
\midrule
\modelname{} & 89.33 $\pm$ 1.15 & 81.82 $\pm$ 0.00 & 0.758 $\pm$ 0.005 \\
CellType & 89.33 $\pm$ 2.31 & 83.33 $\pm$ 6.94 & 0.641 $\pm$ 0.082 \\
Claude Code & 90.00 $\pm$ 2.00 & 83.33 $\pm$ 2.62 & 0.759 $\pm$ 0.012 \\
\bottomrule
\end{tabular}
\end{table}

\subsection{Clinical End-to-End Benchmark}
We constructed a benchmark centered on the translational medicine scenario family, where
each query is paired with expert-authored ground truth. The benchmark tests system’s end-2-end
performance, i.e. whether the system can reason, retrieve evidence, and generalize across a domain
hierarchy. \autoref{tab:main_results}

\begin{table}[!ht]
\caption{End-to-end benchmark results (30 queries): positive hit rate and negative clear rate (\%).
\modelname{}, CellType, Medea, and OpenAI Deep Research: mean~$\pm$~std over 3 runs.
Medea is powered by GPT-5.4 with reasoning effort set to ``low''.}
\label{tab:main_results}
\centering
\small
\begin{tabular}{@{}lcc@{}}
\toprule
\textbf{System} & \textbf{Positive rate} & \textbf{Negative rate} \\
\midrule
\textbf{\modelname{}} & \textbf{\BioEtwoPosMeanPct{} $\pm$ \BioEtwoPosSdPct} & \textbf{\BioEtwoNegMeanPct{} $\pm$ \BioEtwoNegSdPct} \\
CellType & 61.7 $\pm$ 2.3 & 83.3 $\pm$ 0.0 \\
Medea & 33.7 $\pm$ 5.3 & 93.3 $\pm$ 3.3 \\
OpenAI Deep Research & 68.9 $\pm$ 7.7 & 81.1 $\pm$ 1.9 \\
\bottomrule
\end{tabular}
\end{table}

\subsection{Case Study: Generation of Novel Biomarker Hypotheses}
The ATR biomarker discovery case study illustrates our proposed regime in Appendix~\ref{app:cases}.
The system translated a broad synthetic-lethality question into a multi-stage omics analysis, separated pan-cancer from subtype-specific signals, and distinguished mechanistic pathway activation from candidate therapeutic sensitivity. Its final ranking did not simply maximize statistical signal: TP53 loss was treated as a robust marker of ATR pathway activation, ATM loss as the most mechanistically and clinically supported synthetic-lethal biomarker, APC loss as an exploratory colorectal-cancer-specific candidate, and ARID1A loss as biologically plausible but weaker in the available patient data.
This behavior is aligned with the intended use case: producing a reviewable translational dossier that exposes evidence, caveats, and prioritization logic.

%% file: sections_4pages/7_analysis_discussion.tex
\section{Analysis and Discussion}
\label{sec:discussion}

\modelname{} obtains the best overall single-step pass rate (83.49\%) and average judge score (0.892) on the 109-question suite (\autoref{tab:single_step_results}), surpassing the specialized CellType agent (67.80\%, 0.731) and the strongest frontier baseline GPT-5.5 (high) (45.09\%, 0.591).
The largest margin is at L3 (Quantitative Analysis), where \modelname{} reaches 91.45--96.58\% against 46.60\% for CellType and 28.21--30.77\% for GPT-5.5, consistent with the architectural choice to expose genome-wide DepMap-style analyses through a CodeAct sandbox rather than a fixed set of pre-enumerated tool functions.
At the same time, \modelname{} is not uniformly dominant at every layer: CellType achieves a higher L2 pass rate (77.48\% vs.\ 68.47--74.77\%) on qualitative-synthesis questions, and frontier LLMs plateau near a 68\% L1 ceiling regardless of reasoning effort.
For narrow fact retrieval, specialized systems can therefore match or exceed a broader orchestrator.

The open-ended analysis benchmarks reinforce this interpretation.
On BixBench, \modelname{} reaches 89.33\% accuracy on the full set and 81.82\% on the human-biology subset, competitive with package-augmented coding agents (\autoref{tab:bixbench_results}).
On BaisBench Scientific Discovery, \modelname{} achieves a mean score of $0.758 \pm 0.005$, exceeding CellType ($0.641 \pm 0.082$) and matching Claude Code ($0.759 \pm 0.012$; \autoref{tab:baisbench_results}).
These results do not imply that orchestration alone solves autonomous biological discovery.
Rather, they show that a translational agent can retain competitive data-analysis ability while adding the surrounding machinery that scientific users need: entity grounding, scenario-specific methodology, source separation, and final dossier construction.

On the end-to-end benchmark, \modelname{} achieves the highest positive hit rate (\BioEtwoPosMeanPct\%\ $\pm$\ \BioEtwoPosSdPct\%; \autoref{tab:main_results}) and the highest negative clear rate (\BioEtwoNegMeanPct\%\ $\pm$\ \BioEtwoNegSdPct\%), exceeding CellType, Medea, and OpenAI Deep Research on both axes.

%% file: appendix/architecture.tex
\section{Architecture Details}
\label{app:architecture}

\begin{figure}[t]
\centering
\includegraphics[width=\linewidth]{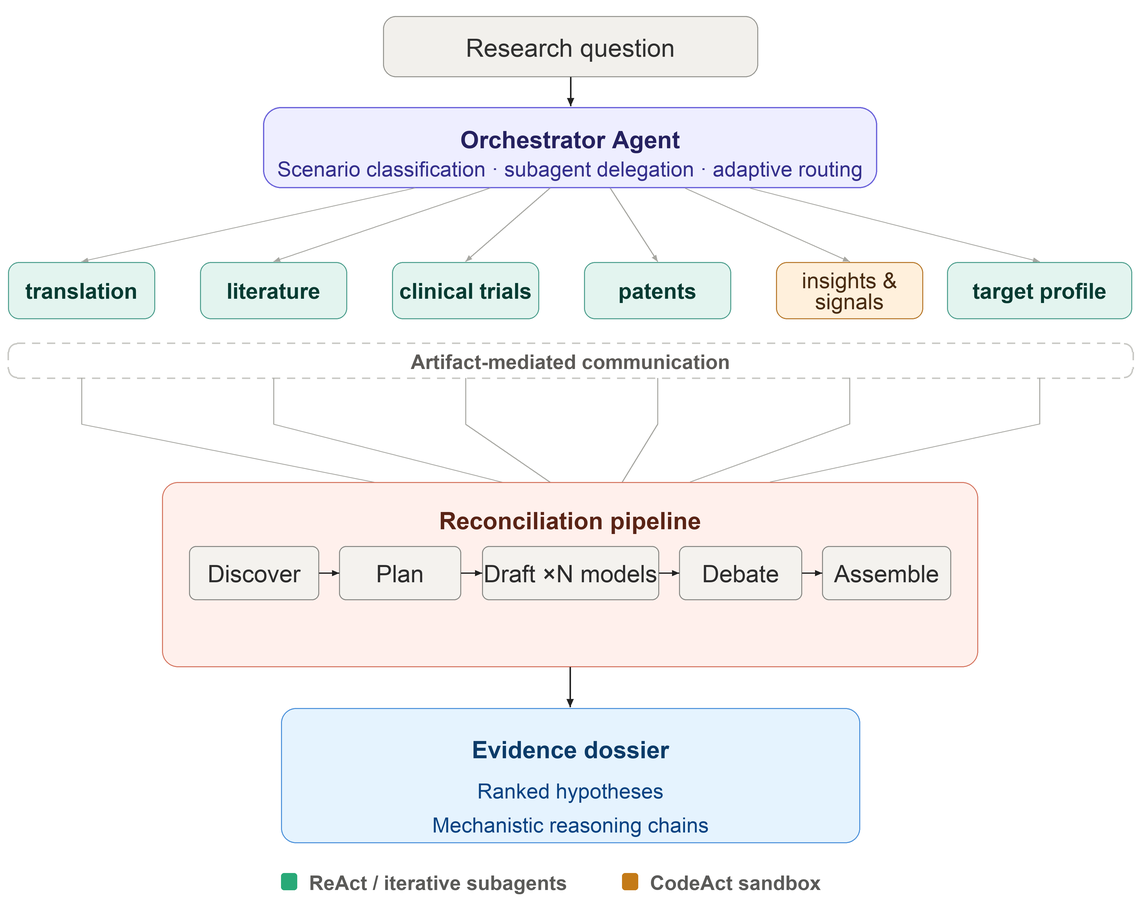}
\caption{System architecture.
The master orchestrator selects a scenario playbook and adaptively delegates self-contained subtasks to specialized retrieval and analysis agents.
Agents publish provenanced source artifacts to a shared evidence bus; a reconciliation agent ingests the bus, drafts the report with multiple models, and resolves disagreement at the claim level.}
\label{fig:architecture}
\end{figure}

\begin{figure}[t]
\centering
\includegraphics[width=\linewidth]{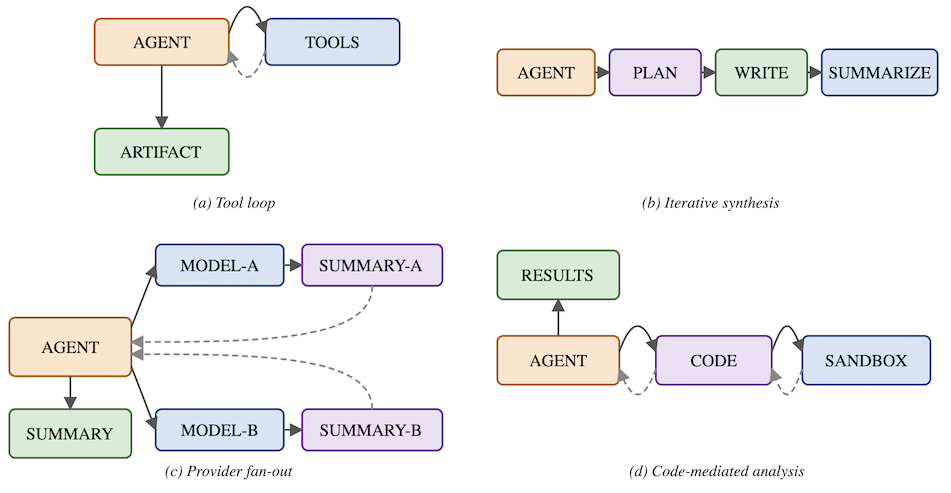}
\caption{Four reusable execution patterns.
\textbf{(a)} Tool loops support structured database access: an agent iterates with a fixed tool surface and emits a single artifact when finished.
\textbf{(b)} Iterative synthesis maps an agent's research output through a deterministic plan, write, and summarize pipeline; the iteration is internal to the agent.
\textbf{(c)} Provider fan-out dispatches the same task to heterogeneous models in parallel; each model produces a summary that the agent composes into a final summary, exploiting cross-family diversity to reduce single-model bias.
\textbf{(d)} Code-mediated analysis lets the agent author code that is executed in a sandbox; sandbox output feeds back into the agent until it emits final results.
Solid arrows: deterministic transitions. Dashed arrows: feedback loops.}
\label{fig:patterns}
\end{figure}

This appendix collects supporting artifacts referenced from Section~\ref{sec:architecture}: the high-level system diagram (\autoref{fig:architecture}), the four reusable execution patterns underlying specialized subagents (\autoref{fig:patterns}), and the claim-level reconciliation procedure (\autoref{alg:debate}).

\begin{algorithm}[t]
\caption{Claim-level section reconciliation.}
\label{alg:debate}
\begin{algorithmic}[1]
\REQUIRE Candidate drafts $D = \{d_1,\ldots,d_M\}$ for section $s$, question $q$, agreement threshold $\tau$, max rounds $R$.
\ENSURE Reconciled section $y_s$.
\STATE For each $d_m$, extract atomic claims with source-support quote and confidence.
\STATE Cluster claims across drafts into topic groups labeled \{full, partial, conflict\}.
\FOR{$r = 1, \ldots, R$}
  \STATE In parallel, each model issues an updated stance per group with revised claim, explanation, and confidence, conditioned on other agents' prior positions.
  \IF{no model disagrees on any group \textbf{and} mean confidence $\geq \tau$}
    \STATE \textbf{break}
  \ENDIF
\ENDFOR
\STATE Synthesize $y_s$ from drafts, claim groups, and final positions, restricted to the section's allocated sources.
\end{algorithmic}
\end{algorithm}

%% file: appendix/tools.tex
\section{Domain-Specific Components and Tool Specifications}
\label{app:tools}

This appendix expands the domain-specific component stack summarized in Section~\ref{sec:architecture} and provides parameter schemas for representative tools from each major category.
All tools are implemented as Python functions with type annotations; a \texttt{@tool} decorator exposes them to agents as structured function-calling tools with auto-generated JSON schemas.

\subsection{Entity Resolution and Biomedical Knowledge Grounding}

The \textbf{translation} subagent is the foundation for all downstream research.
It resolves entity ambiguity by mapping natural-language mentions to canonical identifiers using a ReAct agent with eight tools backed by local SQLite databases (ChEMBL~36, Open Targets) with API fallback:

\begin{itemize}[nosep]
  \item \texttt{target\_name\_to\_chembl\_id}: gene/protein name $\to$ ChEMBL target ID
  \item \texttt{molecule\_name\_to\_chembl\_id}: drug name $\to$ ChEMBL molecule ID
  \item \texttt{target\_to\_drug\_chembl\_ids}: target $\to$ associated drugs (Phase $\geq 2$)
  \item \texttt{compound\_smiles\_to\_name\_and\_chembl\_id}: SMILES $\to$ compound identity
  \item \texttt{get\_drug\_indications}: drug $\to$ approved/investigated indications
  \item \texttt{disease\_name\_to\_efo\_id}: disease name $\to$ EFO/MONDO ontology ID
  \item \texttt{molecule\_smiles\_to\_molecule\_synonyms}: SMILES $\to$ all known synonyms (PubChem)
  \item \texttt{get\_gene\_id\_to\_ensembl\_id\_mapping}: gene symbol $\to$ Ensembl ID
\end{itemize}

Local SQLite is preferred over API access for latency and reproducibility; a swappable backend via \texttt{translation\_factory.py} supports both modes.
The output is a structured entity table with canonical IDs consumed by all downstream subagents, ensuring consistent entity resolution across all evidence streams.

\paragraph{Concrete Example.}
For the query ``\emph{Topo-I inhibitor combinations in NSCLC},'' the translation subagent resolves:
(i)~``Topo-I'' $\to$ gene symbol TOP1 $\to$ ChEMBL target CHEMBL1781 $\to$ Ensembl ENSG00000198900;
(ii)~``NSCLC'' $\to$ EFO\_0003060 (cross-mapped to TCGA ``LUAD''/``LUSC'' and DepMap disease contexts);
(iii)~associated compounds: Irinotecan (CHEMBL481), Topotecan (CHEMBL84), Trastuzumab Deruxtecan (CHEMBL4297564), among 24 Phase~$\geq$2 drugs retrieved via \texttt{target\_to\_drug\_chembl\_ids}.
This single grounding step prevents downstream subagents from searching for ``topoisomerase~I'' in one database and ``TOP1'' in another without recognizing them as equivalent.

\subsection{Tool Ecosystem}

The system integrates 30+ tools spanning six ML endpoints and multiple external data sources (\autoref{tab:tools}).
\autoref{tab:tools} provides a categorized overview.

\begin{table}[!ht]
\caption{Tool ecosystem organized by category, data source, and interaction pattern.
The 30+ tools span structured function-calling (Pattern~1--3 subagents) and CodeAct sandbox functions (Pattern~4).
ML model endpoints provide predictions for drug synergy.}
\label{tab:tools}
\centering
\scriptsize
\begin{tabular}{@{}llll@{}}
\toprule
\textbf{Category} & \textbf{Tools} & \textbf{Data Source} & \textbf{Pattern} \\
\midrule
Entity Resolution & 8 translation tools & ChEMBL~36, Open Targets, PubChem & Local SQLite / REST \\
Literature & \texttt{search\_pubmed}, \texttt{fetch\_abstracts}, etc.\ (4) & PubMed / NCBI E-utilities & REST API \\
Literature & \texttt{get\_target\_disease\_evidence}, etc.\ (4) & Open Targets & GraphQL API \\
Text data RAG & \texttt{search\_conference\_abstracts} (2) & Licenced text data sources & ChromaDB \\
Clinical Trials & \texttt{search\_clinical\_trials} (2) & ClinicalTrials.gov v2 / AACT & REST / SQLite \\
Patents & \texttt{search\_patents} (2) & Google Patents & SerpAPI \\
Target Profile & \texttt{get\_basal\_protein\_info}, etc.\ (4) & DepMap, HPA & Parquet + API \\
Genome-Wide & Dependency, expression, mutation tools (11) & DepMap 24Q4 multi-omics & CodeAct sandbox \\
PPI/Pathways & \texttt{get\_gene\_pair\_annotations}, etc.\ (3) & STRING/Reactome/GO & CodeAct sandbox \\
Drug Synergy & \texttt{predict\_drug\_combination} (1) & DrugComb v2 & HF Gradio API \\
\bottomrule
\end{tabular}
\end{table}

\subsection{Conference Abstract and other licensed data sources - RAG}

Conference abstracts and other licensed data sources contain the earliest reports of clinical results, often 6--12 months before peer-reviewed publication.
This latency gap is critical for translational research.

The indexing pipeline processes conference PDFs via PyMuPDF with conference-specific chunking strategies. Metadata fields are extracted per chunk: conference name, year, page, abstract ID, section header (e.g., ``Breast Cancer: Metastatic''), inferred disease type (mapped to 12 canonical categories including \texttt{nsclc}, \texttt{colorectal\_cancer}, \texttt{melanoma}), gene symbols (regex-matched against 33 key oncology genes including KRAS, EGFR, BRAF, TOP1, BRCA1/2), drug mentions, trial phase (regex-extracted), NCT IDs (pattern \texttt{NCT\textbackslash{}d\{8\}}), and abstract type (late-breaking, oral, poster).
Embeddings are computed via SentenceTransformer (\texttt{all-MiniLM-L6-v2}) and stored in ChromaDB with metadata filtering, enabling queries such as ``\emph{find late-breaking abstracts mentioning TOP1 in NSCLC}.''

\subsection{CodeAct Sandbox for Genome-Wide Analysis}

The \textbf{insights\_and\_signals} subagent operates a sandboxed Python execution environment pre-loaded with nine DepMap 24Q4 Parquet datasets spanning ${\sim}19$K genes $\times$ ${\sim}1{,}600$ cell lines:

\begin{itemize}[nosep]
  \item \textbf{Dependency screens:} CRISPR gene effect (Chronos), CRISPR dependency probability, shRNA gene effect (DEMETER2), shRNA dependency probability
  \item \textbf{Omics:} Gene expression (log$_2$(TPM+1)), copy number (log$_2$(CN ratio+1)), proteomics (normalized abundance), somatic mutations (binary), loss-of-function mutations (binary)
\end{itemize}

\noindent The namespace includes pre-loaded \texttt{pandas}, \texttt{numpy}, and \texttt{scipy.stats}, plus a \texttt{THRESHOLDS} dictionary encoding expert-curated biological cutoffs:
CRISPR likely dependent ($< -0.5$), strongly dependent ($< -1.0$), dependency probability thresholds (dependent $>0.6$, resistant $<0.4$), copy-number gain ($>1.5$), amplification ($>1.9$), loss ($<0.6$), FDR significance ($<0.1$), and minimum sample size ($\geq 3$).
These thresholds ground the LLM's statistical analyses in accepted DepMap conventions, preventing arbitrary cutoff selection.

\paragraph{Security Model.}
\autoref{tab:security} summarizes the sandbox security constraints.

\begin{table}[!ht]
\caption{CodeAct sandbox security model.}
\label{tab:security}
\centering
\small
\begin{tabular}{@{}lp{7cm}@{}}
\toprule
\textbf{Constraint} & \textbf{Details} \\
\midrule
Blocked builtins & \texttt{exec}, \texttt{eval}, \texttt{compile}, \texttt{\_\_import\_\_}, \texttt{open}, \texttt{input}, \texttt{breakpoint} \\
Execution timeout & 600~s per code block \\
Round cap & Maximum 12 code blocks per session \\
Output truncation & 8{,}000 chars (stdout), 4{,}000 chars (stderr) \\
Variable persistence & Only \texttt{msgpack}-serializable primitives (\texttt{str}, \texttt{int}, \texttt{float}, \texttt{bool}, \texttt{list}, \texttt{dict}); DataFrames and numpy arrays excluded \\
Namespace isolation & Filtered \texttt{\_\_builtins\_\_} copy; no filesystem or network access \\
\bottomrule
\end{tabular}
\end{table}

\paragraph{Edge-Case Handling.}
The sandbox addresses three failure modes: (1)~\emph{resource exhaustion}: code exceeding the 600~s timeout or 8{,}000-char output limit is killed mid-execution, and the LLM receives truncated output with a diagnostic message; (2)~\emph{blocked operations}: attempts to call restricted builtins raise \texttt{NameError}, which the LLM can observe and self-correct; (3)~\emph{serialization failures}: variables that cannot be \texttt{msgpack}-serialized (e.g., matplotlib figures) are silently dropped between code blocks, preventing state pollution.

\paragraph{Supported Analyses.}
18 callable tool functions span genome-wide dependency correlations, expression--dependency associations, mutation stratification (Mann--Whitney~$U$), synthetic lethality analysis (absolute thresholds + percentile ranking), PPI network analysis, GEO \citep{clough2024geo} expression outlier detection, and drug combination synergy prediction.

\subsection{Parameter Schemas for Representative Tools}

\paragraph{Patent Search (\texttt{search\_patents}).}
The most complex tool schema, supporting both textual and chemical structure queries:

\begin{table}[!ht]
\centering
\scriptsize
\begin{tabular}{@{}llp{5.5cm}@{}}
\toprule
\textbf{Parameter} & \textbf{Type} & \textbf{Description} \\
\midrule
\texttt{text\_terms} & \texttt{List[TextTerm]?} & Structured query terms. Each \texttt{TextTerm} specifies \texttt{value} (phrase), \texttt{where} $\in$ \{TITLE, ABSTRACT, CLAIM, FULL\_DOCUMENT\}, \texttt{match} $\in$ \{PARTIAL, EXACT\} \\
\texttt{chemical\_terms} & \texttt{List[ChemicalTerm]?} & Chemical structure queries. Each \texttt{ChemicalTerm} specifies \texttt{value} (SMILES or InChIKey), \texttt{molecule\_type} $\in$ \{SMILES, INCHI\_KEY\}, \texttt{chemical\_search\_type} $\in$ \{EXACT, SIMILAR, SUBSTRUCTURE\} \\
\texttt{text\_connector} & \texttt{QueryLogic} & AND/OR for combining text terms \\
\texttt{assignees} & \texttt{List[str]?} & Patent assignee (company) names \\
\texttt{status} & \texttt{StatusTypes} & GRANT / APPLICATION / ALL \\
\texttt{start\_date}, \texttt{end\_date} & \texttt{str?} & Date range (YYYY/MM/DD format) \\
\texttt{limit} & \texttt{int} & Results cap (10--100, default 10) \\
\bottomrule
\end{tabular}
\end{table}

\noindent SMILES inputs are canonicalized via RDKit \texttt{StandardizeSmiles}; InChIKeys are validated by regex.
Backend: SerpAPI \texttt{google\_patents} engine with field-code query construction (e.g., \texttt{TI=} for title, \texttt{SSS=(struct)} for substructure).

\paragraph{Clinical Trial Search (\texttt{search\_clinical\_trials}).}

\begin{table}[!ht]
\centering
\scriptsize
\begin{tabular}{@{}llp{5.5cm}@{}}
\toprule
\textbf{Parameter} & \textbf{Type} & \textbf{Description} \\
\midrule
\texttt{condition} & \texttt{str?} & Disease/condition; supports AND/OR syntax, e.g., \texttt{((NSCLC) OR (lung cancer)) AND (EGFR)} \\
\texttt{intervention} & \texttt{str?} & Drug or treatment name; supports AND/OR \\
\texttt{term} & \texttt{str?} & Free-text keyword search across all study fields \\
\texttt{status} & \texttt{List[str]?} & Phase filter: RECRUITING, COMPLETED, ACTIVE\_NOT\_RECRUITING, etc.\ (14 valid values) \\
\texttt{date\_range} & \texttt{str?} & Primary completion date range, e.g., \texttt{2020-01-01..2023-12-31} \\
\texttt{limit} & \texttt{int} & Max results (default 50, max 1000) \\
\bottomrule
\end{tabular}
\end{table}

\noindent Default returned fields: NCTId, BriefTitle, OverallStatus, BriefSummary, Condition, Phase, InterventionName.
Backend: ClinicalTrials.gov REST API v2.
A local AACT SQLite variant with identical signature is available for offline/reproducible evaluation.

\paragraph{PubMed Search (\texttt{search\_pubmed}).}

\begin{table}[!ht]
\centering
\scriptsize
\begin{tabular}{@{}llp{5.5cm}@{}}
\toprule
\textbf{Parameter} & \textbf{Type} & \textbf{Description} \\
\midrule
\texttt{query} & \texttt{str} & PubMed query using native syntax: field tags (\texttt{[Title/Abstract]}, \texttt{[MeSH Terms]}, \texttt{[Gene]}), Boolean operators, quoted phrases \\
\texttt{max\_results} & \texttt{int} & Result limit (default 80) \\
\texttt{sort} & \texttt{relevance $|$ date} & Sort order \\
\texttt{min\_date}, \texttt{max\_date} & \texttt{str?} & Publication date range (YYYY/MM/DD) \\
\bottomrule
\end{tabular}
\end{table}

\noindent Returns PMID, title, journal, and year per result.
On zero results, the tool returns PubMed's \texttt{querytranslation} and lists unrecognized terms as diagnostic feedback, enabling the agent to reformulate queries.
Backend: NCBI E-utilities (\texttt{esearch} $\to$ \texttt{esummary}) with exponential backoff on rate limiting.

%% file: appendix/benchmarks.tex
\section{Benchmarks Construction}
\label{app:benchmark}

\subsection{Single-Step Tests: Subsets}
\label{app:single_step_subsets}

\begin{table}[!ht]
    \caption{Single-step test subsets, layer coverage, and a representative query per subset.
Each subset targets a specific system layer and is scored independently with its own rubric.}
    \label{tab:single_step_subsets}
    \centering
    \small
    \begin{tabular}{@{}llcp{6.2cm}@{}}
    \toprule
    \textbf{Subset} & \textbf{Layer Tested} & \textbf{N} & \textbf{Representative Query} \\
    \midrule
    Ontology \& Entity & Entity resolution / grounding & 33 & \emph{``What is the Ensembl gene ID and ChEMBL target ID for CD340?''} \\
    Qualitative Synthesis & Literature / trials / patents retrieval & 37 & \emph{``Provide the title of the publication with PMID 25439351.''} \\
    Quantitative Analysis & DepMap / omics computation & 39 & \emph{``What is the CRISPR gene-effect score for PARP1 in MCF7?''} \\
    \midrule
    \textbf{Total} & & \textbf{109} & \\
    \bottomrule
    \end{tabular}
\end{table}

\subsection{Single-Step Tests: Benchmark Design}
\label{app:single_step_design}

The benchmark consists of 109 questions requiring single-step reasoning and is partitioned into three subsets (layers), summarized in \autoref{tab:single_step_subsets}.
\textbf{Ontology \& Entity} (L1) challenges entity normalization across biomedical ontologies.
This layer focuses mostly on the ability of the agent to retrieve key entities (e.g. genes, diseases) from the free-text input and map them to their canonical IDs.
Another example task is resolving a gene family name to a list of member genes with canonical IDs.
\textbf{Qualitative Synthesis} (L2) evaluates retrieval of qualitative knowledge from biomedical databases.
The questions check (among others) for facts about genes, compounds, protein functions and pathways, clinical trials, patents, and literature.
Example tasks are retrieving clinical trial NCT IDs for a given context or assessing the success of a clinical trial given its NCT number.
\textbf{Quantitative Analysis} (L3) assesses extraction and computation of key quantitative data points.
An example task is checking the expression level of a given gene in a given cell line.

\paragraph{Test-case format.}
Each test case is a pair of a question (a natural-language query for the agent) and an expected output (example natural-language response that the agent is expected to return).
For example, the expected output for the question
\emph{``What is the Ensembl gene ID and ChEMBL target ID for CD340?''} is as follows.
\begin{verbatim}
Ensembl ID: ENSG00000141736
ChEMBL target ID: CHEMBL1824
\end{verbatim}
Each expected output was manually verified by an expert to be correct.

\subsection{Single-Step Tests: Evaluation Protocol}
\label{app:single_step_protocol}

\paragraph{Execution.}
Every question is submitted end-to-end to the full master graph: the top-level orchestrator dispatches it to the appropriate sub-agent, which invokes tools as needed.
No sub-agent is tested in isolation, so the single-step tests also probe the orchestrator's routing decisions.
The agent's \emph{actual output} for scoring is the natural-language content of the final message emitted by the master graph.

\paragraph{Judge.}
We use DeepEval~\cite{deepeval}, an LLM-as-a-judge framework, to score open-ended natural-language outputs against the expected outputs.
For each test case, the judge is shown the question, the actual output, and the expected output, and assigns a score between 0 and 1, together with a free-text justification.
Each test case is paired with a custom DeepEval metric that instructs the judge how to score the output.
A metric specification consists of (i) natural-language evaluation criteria, (ii) a rubric, i.e.\ a list of tier descriptions mapped to disjoint score bands on $[0, 1]$, and (iii) a threshold score (the minimum score required for the test to pass).
An example criteria block is:
\begin{verbatim}
Evaluation dimensions:
1. Canonical entity: When the expected output labels a canonical name (e.g. drug name,
disease name), is it present and unambiguously identified in the actual output?
2. Identifier match: Do all identifiers labelled in the expected output appear
in the actual output with matching values?
3. No fabrication: The actual output does not introduce alternative identifiers
that contradict the expected ones, and does not confuse the entity with a same-string
alias of a different gene/disease/drug.
\end{verbatim}
A three-tier rubric yields the bands $[0.0, 0.2]$, $[0.4, 0.6]$, $[0.8, 1.0]$; the gaps make tier boundaries unambiguous and the judge commits to one tier per test case.
In each custom metric, the threshold score is set so that a test passes if the judge assigns a score in the highest band (e.g. 0.8 for three-tier rubrics).
An example three-tier rubric for entity-resolution questions (like the example for layer \emph{Ontology \& Entity} in \autoref{tab:single_step_subsets}) is shown in \autoref{tab:rubric_example}.
We used DeepEval's default LLM model, which is GPT 4.1, for all metrics.

\begin{table}[!ht]
\caption{Example rubric: the judge assigns one tier, yielding a score in the corresponding band.}
\label{tab:rubric_example}
\centering
\small
\begin{tabular}{@{}cp{14cm}@{}}
\toprule
\textbf{Score band} & \textbf{Expected outcome specification} \\
\midrule
$[0.0,\,0.2]$ & Identifiers absent or contradicting the expected ones, or canonical entity (when labelled in the expected output) wrong or missing. \\
$[0.4,\,0.6]$ & Most identifiers correct but some missing or partially mismatched, or canonical entity (when labelled) partially off. Additional information contradicts the expected output. \\
$[0.8,\,1.0]$ & All expected identifiers present and matching, canonical entity (when labelled) correct, no fabricated alternatives. If additional information is present, it doesn't contradict the expected output. \\
\label{app:biomarker-benchmark}
\end{tabular}
\end{table}

\paragraph{Metric.}
The primary per-subset metric is the \emph{pass rate}: the fraction of test cases in that subset whose judge-assigned score satisfies $\text{score} \geq \text{threshold}$
(i.e. such that were classified into the top tier of the rubric).
Another metric is the average metric score: the average score of the test cases.

\subsection{BixBench}
\label{app:bixbench}

We use a subset of BixBench restricted to \textit{Homo sapiens} data, which probes open-ended biomedical data analysis under partial specification.
We report BixBench-style open-answer accuracy on \texttt{phylobio/BixBench-Verified-50}~\cite{mitchener_bixbench_2025} (\autoref{tab:bixbench_results}) for the full 50-question set and a 22-question \emph{human} subset of questions that concern human biology.
For fairness, package-augmented Claude Code variants were provided the same data-analysis package list available to \modelname{}, ensuring comparable minimal execution context.

\subsection{BaisBench}
\label{app:baisbench}

We use the Scientific Discovery (BAIS-SD) track of BaisBench~\cite{luo_benchmarking_2026}:
193 data-driven questions on real single-cell RNA-seq datasets, grounded in biological conclusions from 41 published single-cell studies, where agents analyze the provided data to select answers in single- and multi-answer formats as in the benchmark.
Scoring is trichotomous ($0$, $0.5$, or~$1$) per the SD task:
single-answer items receive $0$ or~$1$; multi-answer items receive $1$ if all correct options are selected and no incorrect one, $0.5$ if at least one correct option is selected and no incorrect one (but not the full set), and~$0$ otherwise.
We report the mean~$S_{\mathrm{SD}}$ in \autoref{tab:baisbench_results}.
\paragraph{Item construction.}
Candidate items were drafted by domain experts and cross-checked against at
least one primary source per item (regulatory document, registered trial with
reported biomarker analysis, or peer-reviewed publication). Items were
retained only if (i)~the positive answer was unambiguously supported by the
primary source and (ii)~the pairing was not trivially derivable from the
query string itself (e.g., we excluded items where the biomarker name
appeared verbatim in the drug's INN or trial title). Tiers~2 and~3 were
deliberately included to stress-test failure modes observed in pilot
evaluations: target/biomarker conflation in Tier~2, and the gap between
mechanistic plausibility and clinical translation in Tier~3.

\subsection{End-2-End Clinical Biomarkers Benchmark Construction}
\paragraph{Scope and design rationale.}
The benchmark targets a single, recurring translational scenario: given a
therapy (approved or investigational), an MOA, or an indication, identify a
biomarker with documented clinical or preclinical relevance. Items were
stratified by evidential maturity so that aggregate scores can be decomposed
into (a)~recall of canonical, label-level facts and (b)~generalization to
settings where signal is distributed across heterogeneous sources (trial
registries, mechanistic literature, functional-genomics screens, observational
prognostic studies). Tier-level expected accuracy was set \emph{a priori}
based on the breadth and consistency of available primary evidence.

\paragraph{Category specification.}
The 30 items are distributed across seven tiers as summarized in
Table~\ref{tab:biomarker-tiers}.

\begin{table}[!ht]
\caption{Tier structure of the translational-medicine biomarker benchmark.
Tier-level expected recall reflects \emph{a priori} difficulty estimates
based on evidential maturity and source breadth.}
\label{tab:biomarker-tiers}
\centering
\small
\renewcommand{\arraystretch}{1.25}
\begin{tabular}{@{}c p{4.8cm} p{3.4cm} c p{1.6cm}@{}}
\toprule
\textbf{Tier} & \textbf{Construct tested} & \textbf{Source archetype} & \textbf{n}  \\
\midrule
1 & Clinically-approved biomarker for an approved therapy (canonical pairing) & FDA/EMA label, NCCN guideline & 4  \\
2 & Clinically-approved biomarker -- testing detailed description & Label $+$ pivotal trial & 4  \\
3 & Synthetic-lethality biomarker with clinical evidence, no full approval & Phase I/II trials, mechanistic papers & 4  \\
4 & Biomarker reported in trials of non-approved therapies & ClinicalTrials.gov, trial publications & 4  \\
5 & Biomarker with extensive preclinical validation (non-approved therapies) & Peer-reviewed preclinical studies & 5   \\
6 & Biomarker from synthetic-lethality discovery studies & CRISPR/RNAi screens, DepMap-class evidence & 5  \\
7 & Prognostic biomarker for a specific indication & Observational and meta-analytic literature & 4 \\
\bottomrule
\end{tabular}
\end{table}

\paragraph{Negative controls.}
For each item, an expert-authored negative biomarker was selected from
adjacent biology---same pathway, same indication class, or same therapeutic
modality---such that surface-level retrieval (keyword overlap, co-occurrence
in literature) would plausibly surface it. This design penalizes systems
that retrieve by topical proximity rather than evidence-grounded reasoning.

\paragraph{Query template.}
Each benchmark item is submitted to the system using the fixed prompt below,
with the placeholders \texttt{[Target (MoA)]} and \texttt{[Indication]}
replaced by the corresponding fields from the item record:
\begin{quote}\small
\textit{Your task is:}

Given a drug with its known molecular target(s): \texttt{[Target (MoA)]}, and an indication \texttt{[Indication]}, return a prioritized list of candidate molecular biomarkers that may predict treatment response, resistance, sensitivity, or patient enrichment.

\textit{Definitions:}
\begin{itemize}[nosep,leftmargin=*]
  \item \emph{Target} --- the molecular entity or pathway directly modulated by the drug.
  \item \emph{Biomarker} --- a molecular feature distinct from the direct drug target, such as mutation, amplification, deletion, fusion, expression change, pathway activation state, homologous recombination deficiency marker, co-mutation pattern, or other molecular alteration associated with response or resistance.
  \item \emph{Indication} --- disease / tumor type / clinical context.
\end{itemize}
\end{quote}

\paragraph{Scoring.}
For each item $i$ with positive answer $p_i$, negative control $n_i$, and a system returning a Top-10 list $L_i$, we report two metrics computed separately and then averaged across all $N$ benchmark items: the \emph{positive sample rate}
\begin{equation}
\mathrm{PSR} \;=\; \frac{1}{N}\sum_{i=1}^{N}\mathbf{1}[p_i \in L_i],
\label{eq:biomarker-psr}
\end{equation}
i.e.\ the fraction of items whose positive answer is recovered, and the \emph{negative sample rate}
\begin{equation}
\mathrm{NSR} \;=\; \frac{1}{N}\sum_{i=1}^{N}\mathbf{1}[n_i \notin L_i],
\label{eq:biomarker-nsr}
\end{equation}
i.e.\ the fraction of items whose negative control is correctly excluded.
The whole benchmark is run three times per system, and the reported PSR and NSR are averaged across the three runs.
Tier-level subscores are reported alongside the aggregate to expose where systems trade off canonical recall against generalization to less mature evidence.

\begin{table}[!ht]
\caption{Full specification of the 30-item translational-medicine biomarker
benchmark (Tiers~1--4). Each item pairs a therapy/MOA/indication with an
expert-validated positive biomarker and an expert-authored negative control
drawn from adjacent biology.}
\label{tab:biomarker-benchmark-full-1}
\centering
\scriptsize
\setlength{\tabcolsep}{4pt}
\renewcommand{\arraystretch}{1.15}
\begin{tabular}{@{}c l p{1.7cm} p{1.4cm} p{1.9cm} p{2.1cm} p{1.1cm} p{2.6cm}@{}}
\toprule
\textbf{\#} & \textbf{Tier} & \textbf{Drug (modality)} & \textbf{Target (MOA)} &
\textbf{Indication} & \textbf{Biomarker (positive)} & \textbf{Negative} &
\textbf{Reference} \\
\midrule
1  & T1 & Trametinib & MEK (MAP2K1, MAP2K2) & Melanoma & BRAF & TP53 & NCT01245062 \\
2  & T1 & Trastuzumab deruxtecan (T-DXd) & ERBB2, TOP1 & Breast, NSCLC, gastric & HER2 & TACSTD2 & NCT03734029; NCT01275677 \\
3  & T1 & Vismodegib & SMO & Basal cell carcinoma; medulloblastoma & PTCH1 & CTNNB1 & NCT00833417 \\
4  & T1 & Palbociclib & CDK4/CDK6 & Advanced breast cancer & ER, HER2 & CCND1 & NCT00721461; NCT01740427 \\
\midrule
5  & T2 & Osimertinib & EGFR & Metastatic \& adjuvant NSCLC & EGFR activating mutations (ex19del, L858R; T790M for 2L) & BRCA1 & NCT02296125 \\
6  & T2 & Erdafitinib & FGFR1/2/3/4 & Metastatic urothelial carcinoma & FGFR3 mutations (R248C, S249C, G370C, Y373C); FGFR3--TACC3 fusion & CD274 & NCT03390504 \\
7  & T2 & Mirvetuximab soravtansine & FOLR1, TUBB & Ovarian cancer & FR$\alpha$ IHC $\geq$75\% cells at $\geq$2+/3+ & BRCA1/2 & NCT02631876; NCT04296890 \\
8  & T2 & Elacestrant & ESR1 & Advanced breast cancer & \emph{ESR1} LBD missense mutations (ctDNA) & ER & NCT03778931 \\
\midrule
9  & T3 & Entrectinib & ROS1 & Lobular breast, gastric, TNBC & CDH1 & RHOA & NCT04551495; NCT03620643 \\
10 & T3 & CPI-0209 & EZH2 & Advanced solid tumours \& lymphomas & ARID1A & KMT2D & NCT04104776 \\
11 & T3 & RP-6306 & PKMYT1 & Advanced solid tumours & FBXW7 & TP53 & NCT04855656 \\
12 & T3 & Onvansertib & PLK1 & Metastatic CRC & KRAS & APC & NCT03829410 \\
\midrule
13 & T4 & Everolimus & mTOR & Thyroid cancer & TSC1 & ESR1 & PMID:25295501 \\
14 & T4 & Olaparib & PARP1, PARP2 & Prostate cancer & BRIP1 & PARG & NCT02987543 \\
15 & T4 & Vemurafenib & BRAF & Pancreatic cancer & KRAS & PTEN & NCT01524978 \\
16 & T4 & Palbociclib & CDK4/6 & Myeloma & CDKN2A & MDM2 & NCT00555906 \\
\bottomrule
\end{tabular}
\end{table}

\begin{table}[!ht]
\caption{Full specification of the 30-item translational-medicine biomarker
benchmark (Tiers~5--7), continued from Table~\ref{tab:biomarker-benchmark-full-1}.}
\label{tab:biomarker-benchmark-full-2}
\centering
\scriptsize
\setlength{\tabcolsep}{4pt}
\renewcommand{\arraystretch}{1.15}
\begin{tabular}{@{}c l p{1.7cm} p{1.4cm} p{1.9cm} p{2.1cm} p{1.1cm} p{2.6cm}@{}}
\toprule
\textbf{\#} & \textbf{Tier} & \textbf{Drug (modality)} & \textbf{Target (MOA)} &
\textbf{Indication} & \textbf{Biomarker (positive)} & \textbf{Negative} &
\textbf{Reference} \\
\midrule
17 & T5 & --- & BRAF & Melanoma & MAP2K1 & BRCA1/2 & NCT01006980 \\
18 & T5 & --- & HSP90 & Acute lymphoblastic leukemia & JAK1 & CDKN2A & PMID:22271575 \\
19 & T5 & --- & AURKA & NSCLC & SMARCA4 & TP53 & PMID:28102363 \\
20 & T5 & --- & CDK1/2/5/9 & Osteosarcoma & CCNE1 & VEGFA & PMID:30266815 \\
21 & T5 & --- & FANCM & Pan-cancer & SMARCAL1 & POLE2 & PMID:39510066 \\
\midrule
22 & T6 & --- & PRKDC & Pan-cancer & ATM & POLR2G & PMID:23761041 \\
23 & T6 & --- & PIK3CB & Pan-cancer & PTEN & ARID1A & PMID:18594509 \\
24 & T6 & --- & ROCK2 & Pan-cancer & BRCA2 & IDH1 & PMID:37073955 \\
25 & T6 & --- & LIG1 & Pan-cancer & BRCA2 & POLD1 & PMID:41070620 \\
26 & T6 & --- & FEN1 & Pan-cancer & WDR48 & RUVBL2 & PMID:40205037 \\
\midrule
27 & T7 & --- & --- & Myelodysplastic syndromes & NRAS & PTEN & PMID:38319256 \\
28 & T7 & --- & --- & Colorectal cancer & POLD1 & ESR1 & PMID:38777726 \\
29 & T7 & --- & --- & Invasive breast carcinoma & CCND1 & IDH1 & PMID:18823530 \\
30 & T7 & --- & --- & Medulloblastoma & CTNNB1 & ERBB2 & NCT01878617 \\
\bottomrule
\end{tabular}
\end{table}

\clearpage

%% file: appendix/related_work.tex
\section{Related Work}
\label{app:related}

\subsection{Agentic Systems in Biomedical Research}

LLM-based agent systems in the life sciences span autonomous experimental design, clinical decision support, and broader biological investigation.
In experimental design, Coscientist~\cite{boiko_autonomous_2023} and ChemCrow~\cite{m_bran_augmenting_2024} demonstrated LLM-driven robotic synthesis and tool-grounded chemical reasoning, with later work extending to retrosynthesis~\cite{kang_retrointext_2024}, in-silico lead generation~\cite{averly_liddia_2025}, protein binder design~\cite{pacesa_one-shot_2025}, multi-objective protein engineering~\cite{zhu_advancing_2026, ponnapati_proteincrow_2025, roohani_biodiscoveryagent_2024, ghafarollahi_protagents_2024}, and early clinical validation of an AI-designed kinase inhibitor~\cite{xu_generative_2025}.
These systems typically address well-scoped molecular or synthetic design tasks rather than multi-source evidence integration.

Clinical agents have been applied to therapeutic question answering over curated knowledge bases~\cite{gao_txagent_2025}, adaptive multi-agent triage~\cite{kim_mdagents_2024}, multimodal precision oncology~\cite{ferber_development_2025}, patient-to-trial matching~\cite{jin_matching_2024}, and histopathology pipeline construction~\cite{vaidya_nova_2025}.
They operate within structured workflows such as diagnosis, treatment selection, and eligibility screening, but are not primarily designed for open-ended translational evidence synthesis across heterogeneous sources.

Broader biological agent platforms include Virtual Lab~\cite{virtuallab2024}, Biomni~\cite{huang_biomni_2025}, BioLab~\cite{jin_biolab_2025}, OriGene~\cite{zhang_origene_2025}, and Medea~\cite{sui_medea_2026}, alongside systems for bioinformatics workflow automation~\cite{su_biomaster_2025, pickard_language_2024}, self-verified gene-set analysis~\cite{wang_geneagent_2025}, and knowledge-graph construction~\cite{lu_karma_2025}.
Our system falls within this category but focuses on translational drug-discovery use cases, integrating evidence from literature, clinical trials, patents, and multi-omics datasets into structured research dossiers for preclinical and clinical development decisions.

%% file: appendix/cases.tex
\clearpage

\section{Case Study: Generation of Novel Biomarker Hypotheses using Synthetic Lethality Concept and Patient Data}
\label{app:cases}
\label{sec:cases}

We evaluated the TRESR trial~\cite{yap_camonsertib_2023} (NCT04497116), which is based on a synthetic-lethality (SL) hypothesis linking ATR inhibition to selected loss-of-function (LoF) alterations.
While the study refers to ``ATR inhibitor--sensitizing mutations,'' these are not explicitly defined, though subsequent reports suggest a focus on DNA damage response genes identified via chemogenomic CRISPR screens [PMID: 37277454].

Thus, we tasked \textsc{\modelname{}} with generating SL biomarker hypotheses for ATR inhibition based on clinical data (input specification in \autoref{fig:case_input}).
The agent autonomously selected TCGA and METAPRISM cohorts, identified LoF events associated with ATR expression, and evaluated candidates through survival analyses.
Specifically, the system executed an end-to-end translational pipeline within the multi-agent orchestration framework (total runtime $\approx$1352 s; see \autoref{fig:case_pipeline} for the full pipeline view).
Scenario classification routed the query to the \texttt{combination\_discovery} workflow, after which the \texttt{translation} subagent canonicalized ATR and the relevant clinical indications.
Subsequently, the \texttt{insights\_and\_signals} (CodeAct) subagent conducted downstream multi-omic and survival analyses across multiple cohorts, leveraging iterative model invocations and tool-mediated operations to complete the analysis under the system-generated methodology summarized in \autoref{fig:case_methodology}.

The \textsc{\modelname{}} output combined patient-level data with literature and clinical context, evaluating two signal types: (i) LoF-associated ATR upregulation and (ii) survival patterns in low-ATR tumors as a proxy for drug-induced ATR inhibition.
A condensed reasoning trace integrating these signals into a candidate ranking is shown in \autoref{fig:case_reasoning}.

Across both cohorts, increased ATR expression was most robustly associated with TP53 loss, suggesting a pan-cancer marker of ATR pathway activation, though likely reflecting general replication stress rather than specific therapeutic sensitivity.
In contrast, the strongest evidence for a synthetic-lethal interaction emerged from ATM loss, which - despite lacking compensatory ATR upregulation - showed the clearest survival signal in low-ATR tumors, consistent with prior preclinical and clinical findings.

Subtype-specific signals further highlighted APC loss in colorectal cancer as the most prominent data-driven discovery, supported by both elevated ATR expression and favorable low-ATR survival patterns, although it lacks prior clinical validation.
Finally, ARID1A loss demonstrated only modest support in patient datasets but remains biologically plausible given consistent evidence from external studies, warranting consideration as a secondary translational candidate.
The agent-generated biological interpretation across these candidates is summarized in \autoref{fig:case_interpretation}.

Overall, the integrated evidence supports the following ranked hypotheses: TP53 LoF as the strongest pan-cancer marker of ATR pathway activation; ATM LoF as the most clinically and mechanistically supported synthetic-lethal biomarker; APC LoF in colorectal cancer as an unexpected data-driven candidate; and ARID1A LoF as a literature-supported but modestly recovered clinical candidate (clinically actionable prioritization in \autoref{fig:case_prioritization}).
The framework's key strength lies in combining autonomous multi-cohort analysis with calibrated restraint: where survival models were sparse or unstable, the agent avoided over-interpretation, preserving the reliability of the resulting translational dossier.

The remainder of this appendix shows the artifacts produced at each stage of the pipeline.

\clearpage

\begin{center}
\includegraphics[width=\textwidth,height=0.82\textheight,keepaspectratio]{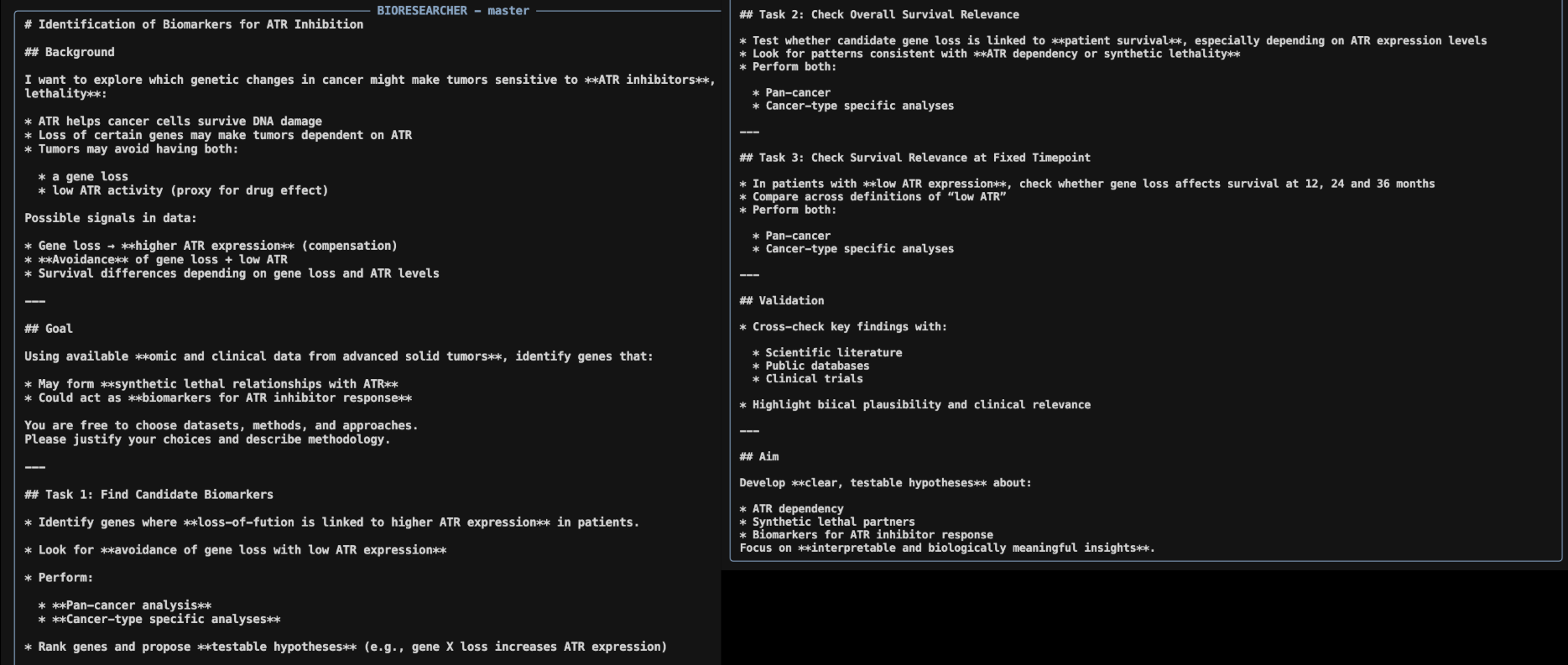}
\captionof{figure}{\textbf{Research goals for automated ATR inhibitor biomarker discovery.}
Input specification to the \modelname{} system defining the biological context (ATR synthetic lethality), expected signals (e.g., gene loss linked to high ATR expression, depletion in low-ATR tumors, survival patterns), and analysis workflow.
The pipeline comprises three tasks: (1) candidate biomarker identification via pan-cancer and subtype-specific loss-of-function and ATR expression analyses; (2) survival association conditioned on ATR levels; and (3) fixed-timepoint survival analysis in low-ATR subgroups.
A final validation step cross-references results with literature, databases, and clinical trials to ensure biological and clinical relevance.
The specification acts as a programmatic interface linking user intent to agent execution, enabling reproducible and interpretable biomarker discovery while constraining the search space.}
\label{fig:case_input}
\end{center}

\clearpage

\begin{center}
\includegraphics[width=\textwidth,height=0.82\textheight,keepaspectratio]{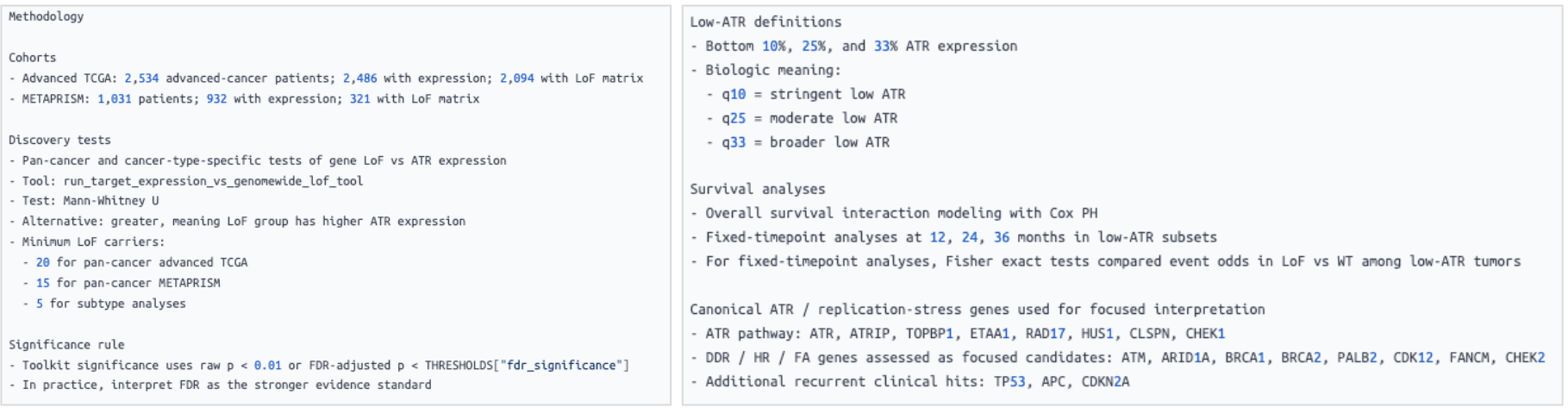}
\captionof{figure}{\textbf{System-generated methodology for automated ATR inhibitor biomarker discovery.}
Methodological specification produced by \modelname{}.
Cohort composition (TCGA, METAPRISM) and sample sizes for expression and LoF data is shown.
Discovery is defined as genome-wide testing of LoF vs.\ ATR expression using directional Mann--Whitney $U$ tests, with cohort-specific carrier thresholds and significance assessed via $p$-values and FDR (primary criterion).
Low-ATR subgroups are defined by expression quantiles (bottom 10\%, 25\%, 33\%) for downstream analyses.
Survival analysis combines Cox interaction models with fixed-timepoint analyses (12, 24, 36 months), supplemented by Fisher's exact tests in sparse settings.
Results are contextualized using canonical ATR and DNA damage response genes.
The framework exposes all analytical choices, ensuring reproducibility and interpretability.}
\label{fig:case_methodology}
\end{center}

\clearpage

\begin{center}
\includegraphics[width=\textwidth,height=0.82\textheight,keepaspectratio]{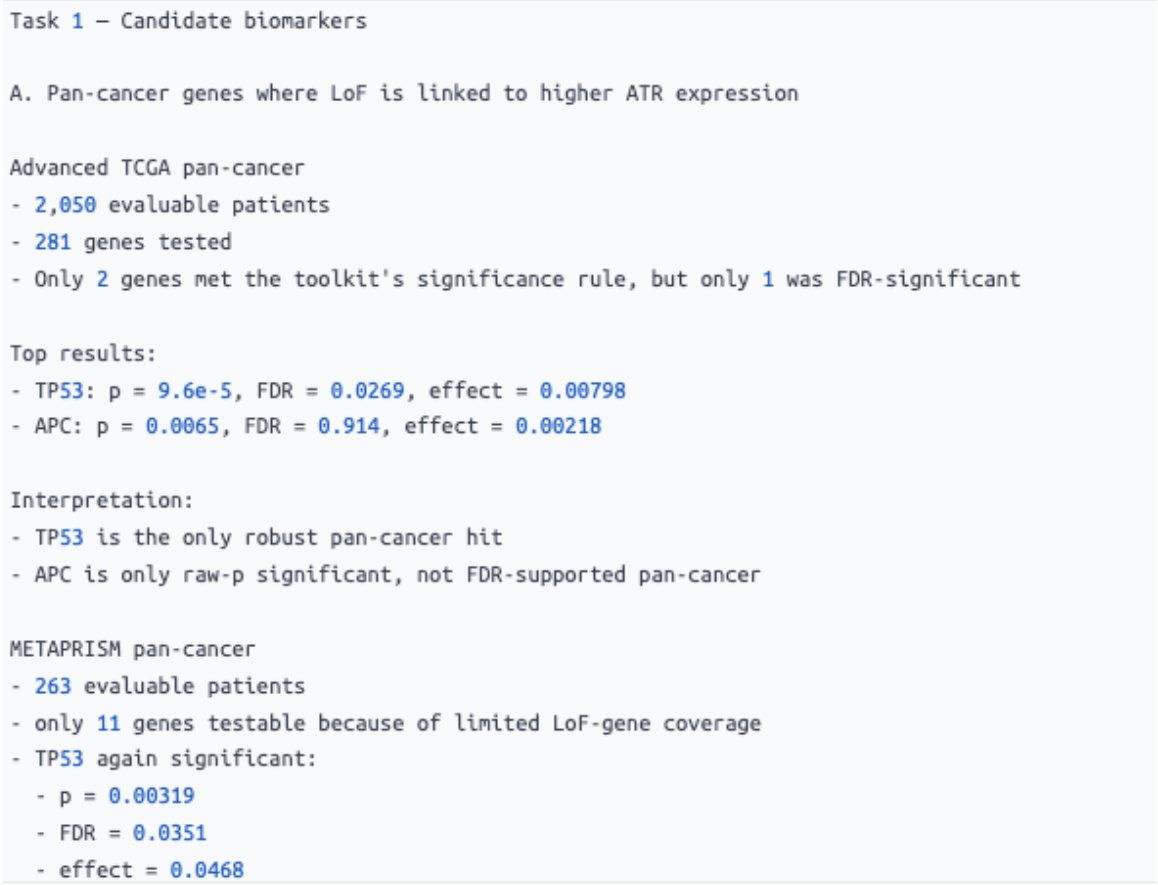}
\captionof{figure}{\textbf{Multi-signal biomarker prioritization from agent reasoning trace.}
Condensed execution trace highlighting how \modelname{} integrates heterogeneous evidence to prioritize ATR biomarker candidates.
TP53 loss emerges as the only reproducible pan-cancer signal associated with increased ATR expression across cohorts.
Subtype-specific analysis identifies APC loss as a strong colorectal cancer signal.
In contrast, survival-based analyses reveal ATM loss as the most synthetic-lethality-like candidate, showing improved outcomes in low-ATR tumors despite lacking expression association.
The system explicitly downweights unstable Cox interaction models and deprioritizes inconsistent signals (e.g., CDKN2A), illustrating calibrated uncertainty handling.
This figure demonstrates cross-task synthesis, where expression, stratified survival, and cohort-specific analyses jointly inform hypothesis ranking.}
\label{fig:case_reasoning}
\end{center}

\clearpage

\begin{center}
\includegraphics[width=\textwidth,height=0.82\textheight,keepaspectratio]{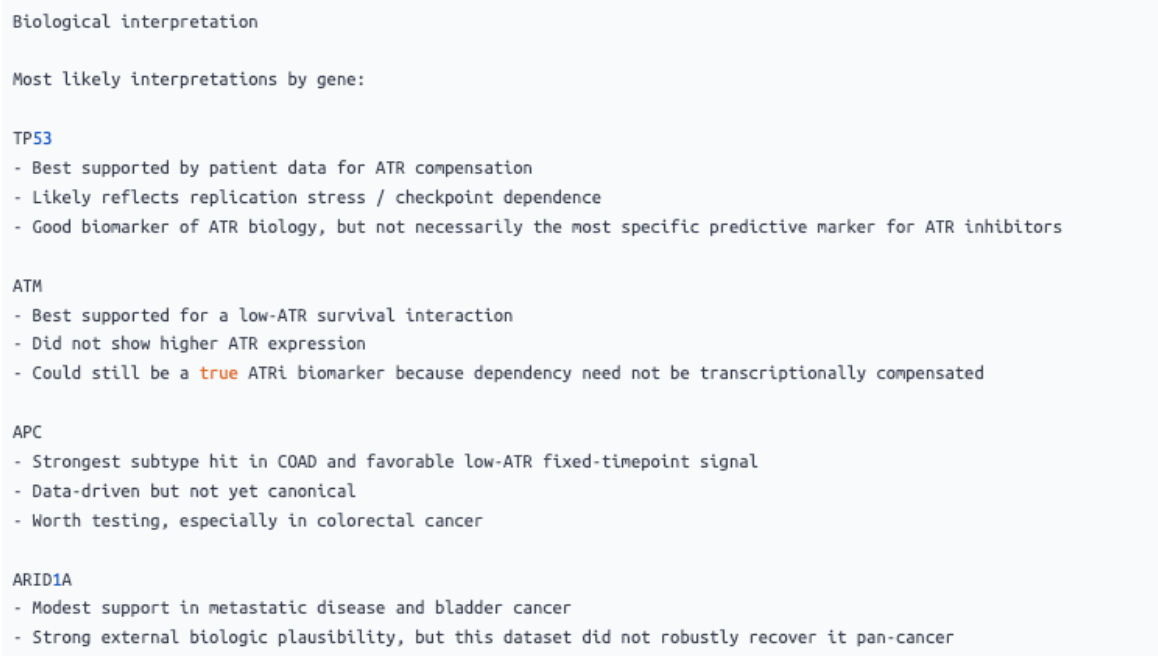}
\captionof{figure}{\textbf{Agent-generated biological interpretation and hypothesis prioritization.}
Post-analysis interpretation of candidate ATR biomarkers, summarizing gene-specific evidence across expression and survival signals.
The system distinguishes between robust, data-supported signals (e.g.\ TP53, ATM) and exploratory or context-specific candidates (e.g.\ APC, ARID1A), producing biologically grounded, testable hypotheses.}
\label{fig:case_interpretation}
\end{center}

\clearpage

\begin{center}
\includegraphics[width=\textwidth,height=0.82\textheight,keepaspectratio]{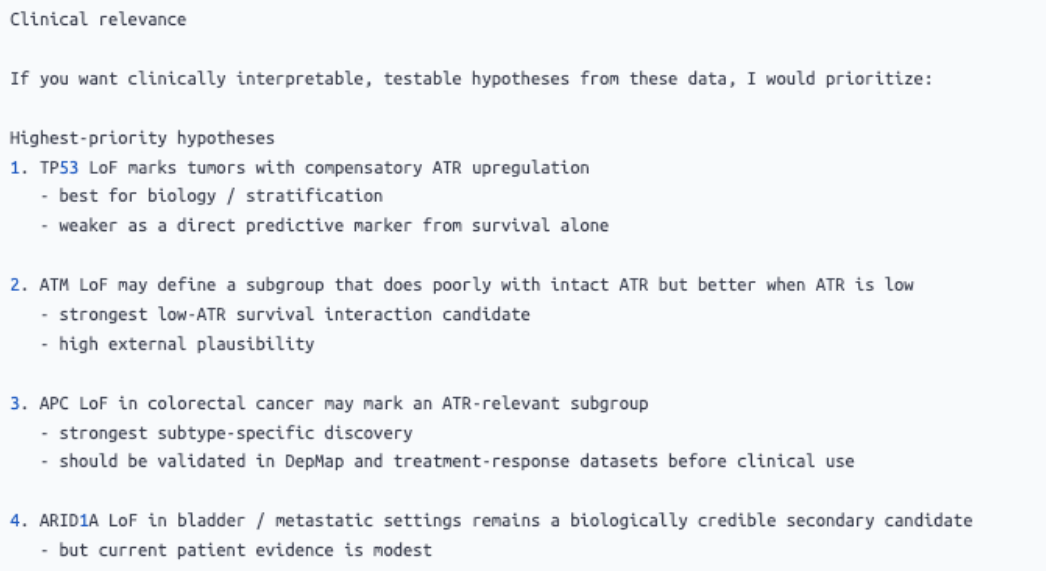}
\captionof{figure}{\textbf{Translation of hypotheses into clinically actionable prioritization.}
Agent-generated ranking of ATR biomarker hypotheses based on clinical interpretability, integrating expression, survival, and external plausibility signals.
The system distinguishes between stratification markers (e.g.\ TP53), synthetic-lethal candidates (e.g.\ ATM), and exploratory, subtype-specific hypotheses (e.g.\ APC, ARID1A).}
\label{fig:case_prioritization}
\end{center}

\clearpage

\begin{center}
\includegraphics[width=\textwidth,height=0.82\textheight,keepaspectratio]{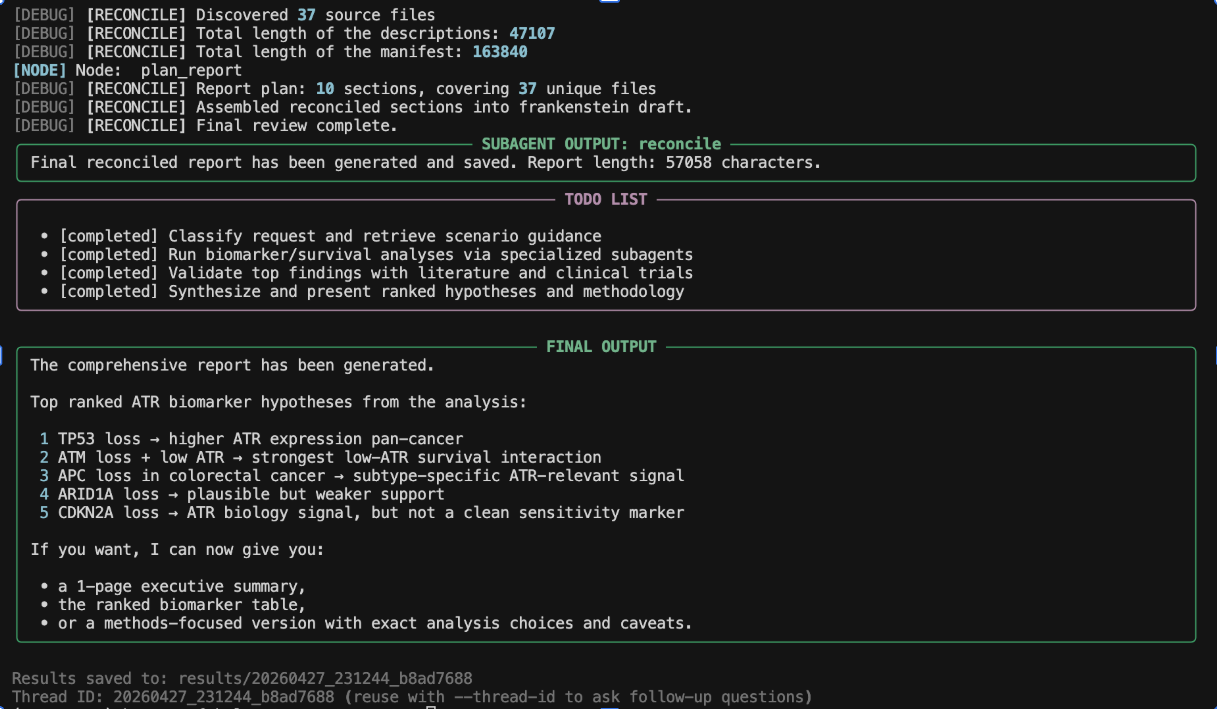}
\captionof{figure}{\textbf{End-to-end execution and hypothesis synthesis in \modelname{}.}
Multi-agent pipeline for ATR biomarker discovery comprising four stages: (1) task decomposition, (2) specialized biomarker and survival analyses, (3) validation with literature and clinical trials, and (4) reconciliation and synthesis.
The system integrates 37 intermediate artifacts into a single report ($\sim$57k characters), completing all steps autonomously.
Final outputs are ranked, testable hypotheses, including TP53 loss (pan-cancer ATR activation), ATM loss (top synthetic-lethal candidate), APC loss (colorectal-specific), and ARID1A loss (weaker signal), while inconsistent findings (e.g., CDKN2A) are deprioritized.
The framework demonstrates end-to-end automation, large-scale result synthesis, and calibrated uncertainty handling.}
\label{fig:case_pipeline}
\end{center}

\clearpage

%% file: appendix/licences.tex
\section{Asset Provenance, Licenses, and Terms of Use}
\label{app:licenses}

This appendix consolidates the third-party assets used by \textsc{BioResearcher}
and in our evaluation. For each asset we report the version (where applicable),
the original creator or maintainer, a citation, a URL, the license under which
the asset is made available, and a brief note on how we use it. Asset usage was
verified against each license at the time of submission.


\begin{table}[p]
  \centering
  \small
  \begin{tabular}{p{2.8cm} p{1.4cm} p{2.6cm} p{2.4cm} p{2.6cm} p{2.6cm}}
    \toprule
    \textbf{Asset} & \textbf{Version} & \textbf{Owner / Maintainer} &
    \textbf{Citation \& URL} & \textbf{License / Terms} & \textbf{Use in this work} \\
    \midrule
    \multicolumn{6}{l}{\textit{(a) Biomedical databases and ontologies}} \\
    \midrule
    ChEMBL & 36 & EMBL-EBI &
    \url{https://www.ebi.ac.uk/chembl/} &
    CC-BY; EBI Terms of Use & Local SQLite; entity resolution and
    target/drug lookup. \\

    Open Targets Platform & 26.03 & Open Targets / EMBL-EBI &
    \url{https://platform.opentargets.org/} &
    Terms of use for the Open Targets Platform - updated June 2021 & Local SQLite + GraphQL API; target--disease
    evidence. \\

    PubChem & 2026-04-29 & NCBI / NLM &
    \url{https://pubchem.ncbi.nlm.nih.gov/} &
    NCBI/NLM public domain; NCBI Website and Data Usage Policies & Compound
    synonym resolution from SMILES. \\

    Ensembl & Release 115 & EMBL-EBI &
    \url{https://www.ensembl.org/} &
    Ensembl Privacy Notice v 3.0.0 & Gene-symbol to Ensembl-ID mapping. \\

    EFO &  & EMBL-EBI &
    \url{https://www.ebi.ac.uk/efo/} &
    Licensing of EMBL-EBI data resources & Disease-name to ontology-ID resolution. \\

    MONDO & v2026-04-07 & Monarch Initiative &
    \url{https://mondo.monarchinitiative.org/} &
    CC~BY~4.0 & Cross-mapped disease IDs. \\

    DOID &  & Disease Ontology consortium &
    \url{https://disease-ontology.org/} &
     & Cross-mapped disease IDs. \\

    Human Protein Atlas & 25.0 & SciLifeLab / KTH &
    \url{https://www.proteinatlas.org/} &
    CC~BY-SA~4.0 &  Basal protein characteristics and for Uniprot ID mapping.\\

    STRING & 12.0 & SIB / EMBL &
    \url{https://string-db.org/} &
    CC~BY~4.0 & Protein--protein interaction graph. \\

    Reactome & v96 & Reactome consortium &
    \url{https://reactome.org/} &
    CC~BY~4.0 & Pathway annotations. \\

    GEO (via ARCHS4 pipeline) &
    ARCHS4 pipeline Apache 2.0 &
    NCBI / NLM &
    \url{https://www.ncbi.nlm.nih.gov/geo/} &
    GEO data: public domain; ARCHS4 pipeline: Apache~2.0 &
    GEO expression outlier detection \\

    Gene Ontology & 2026-03-25 & GO Consortium &
    \url{http://geneontology.org/} &
    CC~BY~4.0 & Functional annotations. \\

    DrugComb & 2 & Changzhou University and CSU &
    \url{http://drugcombdb.denglab.org/} &
     & Drug-pair synergy prediction endpoint. \\

    ClinicalTrials.gov / AACT &  & NLM / CTTI &
    \url{https://aact.ctti-clinicaltrials.org/} &
    AACT user agreement & Trial registration data. \\

    PubMed & live API & NCBI / NLM &
    \url{https://www.ncbi.nlm.nih.gov/} &
    NCBI/NLM Disclaimer & Literature retrieval. \\

    Google Patents (SerpAPI) & live API & SerpAPI (third-party) &
    \url{https://serpapi.com/google-patents-api} &
    SerpAPI Terms of Service (commercial API) & Patent search. \\
    \bottomrule
  \end{tabular}
  \caption{Third-party assets used in \textsc{BioResearcher} and in evaluation.}
  \label{tab:licenses}
\end{table}

\begin{table}[p]
  \centering
  \small
  \caption*{Table~\ref{tab:licenses} (\textit{continued}).}
  \begin{tabular}{p{2.8cm} p{1.4cm} p{2.6cm} p{2.4cm} p{2.6cm} p{2.6cm}}
    \toprule
    \textbf{Asset} & \textbf{Version} & \textbf{Owner / Maintainer} &
    \textbf{Citation \& URL} & \textbf{License / Terms} & \textbf{Use in this work} \\
    \midrule
    \multicolumn{6}{l}{\textit{(b) Functional-genomics and patient cohorts}} \\
    \midrule
    DepMap (CRISPR, shRNA, omics, proteomics, mutations) & 25Q3 &
    Broad Institute &
    \url{https://depmap.org/portal/} & DepMap Terms of Use
    & Sandboxed Python analyses (dependency, expression, mutation
    stratification).  \\

    TCGA & 45.0 &
    NCI / NHGRI &
    \url{https://portal.gdc.cancer.gov/} &
    NIH Genomic Data Sharing Policy &
    Survival analyses; ATR case study. We do not redistribute controlled-access
    data. \\

    MSK-CHORD & 2024 &
    MSK Cancer Data Science Initiative Group &
    \url{https://www.cbioportal.org/study/summary?id=msk_chord_2024} &
    BY-NC-ND 4.0 &
    Survival analyses; ATR case study. \\

    METAPRISM & 30/10/2022 &  European Bioinformatics Institute (EMBL-EBI) &
    \url{https://www.ega-archive.org/} & &
    Survival analyses; ATR case study. \\
    \bottomrule
  \end{tabular}
\end{table}

\begin{table}[p]
  \centering
  \small
  \caption*{Table~\ref{tab:licenses} (\textit{continued}).}
  \begin{tabular}{p{2.8cm} p{1.4cm} p{2.6cm} p{2.4cm} p{2.6cm} p{2.6cm}}
    \toprule
    \textbf{Asset} & \textbf{Version} & \textbf{Owner / Maintainer} &
    \textbf{Citation \& URL} & \textbf{License / Terms} & \textbf{Use in this work} \\
    \midrule
    \multicolumn{6}{l}{\textit{(d) Foundation models, agents, and APIs}} \\
    \midrule
    GPT-5.4-mini, GPT-5.4, GPT-5.5 & API: 2.14.0 & OpenAI &
    \url{https://platform.openai.com/} & OpenAI Business Terms; usage policies &
    Core models for \textsc{BioResearcher} and baselines. \\

    GPT-4.1 & API: 2.14.0 & OpenAI &
    \url{https://platform.openai.com/} & OpenAI Business Terms &
    LLM-as-a-judge for single-step tests. \\

    OpenAI Codex (CLI) & 0.128.0 & OpenAI &
    \url{https://openai.com/} & OpenAI Business Terms & Baseline. \\

    OpenAI Deep Research & API: 2.14.0 & OpenAI &
    \url{https://openai.com/} & OpenAI Business Terms & Baseline. \\

    Gemini 3.1 Pro & API: 1.73.0 & Google &
    \url{https://ai.google.dev/} & Google APIs Terms of Service &
    Baseline. \\

    Claude Code & 2.1.128 & Anthropic &
    \url{https://www.anthropic.com/} & Anthropic Commercial Terms of Service &
    Baseline. \\

    CellType Agent & 28-03-2026 & Jiawen Chen, Jianghao Zhang, Huaxiu Yao, Yun Li &
    \url{https://github.com/jianghao-zhang/CellTypeAgent/} & & Baseline. \\

    Medea & 26-03-2026 & \cite{sui_medea_2026} &
    \url{https://github.com/mims-harvard/MEDEA} & Apache-2.0 & Baseline. \\

    all-MiniLM-L6-v2 & 2.0 & SentenceTransformers &
    \url{https://huggingface.co/sentence-transformers/all-MiniLM-L6-v2} &
    Apache~2.0 & Embedding model for the conference RAG index. \\
    \bottomrule
  \end{tabular}
\end{table}

\begin{table}[p]
  \centering
  \small
  \caption*{Table~\ref{tab:licenses} (\textit{continued}).}
  \begin{tabular}{p{2.8cm} p{1.4cm} p{2.6cm} p{2.4cm} p{2.6cm} p{2.6cm}}
    \toprule
    \textbf{Asset} & \textbf{Version} & \textbf{Owner / Maintainer} &
    \textbf{Citation \& URL} & \textbf{License / Terms} & \textbf{Use in this work} \\
    \midrule
    \multicolumn{6}{l}{\textit{(e) Software libraries and frameworks}} \\
    \midrule
    DeepEval & 3.9.7 & Confident AI & \cite{deepeval};
    \url{https://github.com/confident-ai/deepeval} & Apache~2.0 &
    LLM-as-a-judge framework. \\

    ChromaDB & 1.5.7 & Chroma &
    \url{https://github.com/chroma-core/chroma} & Apache~2.0 &
    Vector store for the conference RAG index. \\

    RDKit & 2026.3.1 & RDKit consortium &
    \url{https://www.rdkit.org/} & BSD-3-Clause & SMILES standardisation. \\

    pandas, numpy, scipy & 2.3.3 / 2.1.3 / 1.17.1 & NumFOCUS &
    \url{https://numpy.org/}, etc. & BSD-3-Clause & Sandbox numerics. \\

    LangGraph / LangChain & 1.1.6 / 1.2.15 & LangChain Inc. &
    \url{https://github.com/langchain-ai/langgraph} & MIT &
    Orchestration framework for the agent graphs. \\
    \bottomrule
  \end{tabular}
\end{table}

\begin{table}[p]
  \centering
  \small
  \caption*{Table~\ref{tab:licenses} (\textit{continued}).}
  \begin{tabular}{p{2.8cm} p{1.4cm} p{2.6cm} p{2.4cm} p{2.6cm} p{2.6cm}}
    \toprule
    \textbf{Asset} & \textbf{Version} & \textbf{Owner / Maintainer} &
    \textbf{Citation \& URL} & \textbf{License / Terms} & \textbf{Use in this work} \\
    \midrule
    \multicolumn{6}{l}{\textit{(f) Benchmarks}} \\
    \midrule
    BixBench-Verified-50 & 50-question subset & FutureHouse / Phylobio &
    \cite{mitchener_bixbench_2025};
    \url{https://huggingface.co/datasets/phylobio/BixBench-Verified-50} &
    Apache~2.0 & Quantitative-reasoning evaluation. \\

    BaisBench (Scientific Discovery, BAIS-SD) & 193-question track &
    Luo et al. & \cite{luo_benchmarking_2026};
    \url{https://github.com/EperLuo/BaisBench} & MIT &
    Open-ended biological-discovery evaluation. \\
    \bottomrule
  \end{tabular}
\end{table}

\paragraph{DepMap compliance note.}
The Broad Institute DepMap Terms of Use permit AI/ML model use with or on the
data \emph{for internal research use, including non-profit sharing of
methodologies}, but restrict commercial use and explicitly restrict use of the
data to ``train, develop, or enhance machine learning or AI models other than
for internal research use.''
\textsc{BioResearcher}'s sandbox loads DepMap~24Q4 Parquet files for
on-the-fly statistical analysis (correlations, Mann--Whitney tests, percentile
ranking, survival analysis). DepMap data are \emph{not} used to fine-tune any
model in this work, are not redistributed in our released artifacts, and are
re-fetched by the user from the official DepMap portal when reproducing our
results.